\documentclass[10pt,journal]{IEEEtran}
\usepackage{amsmath,amssymb,amsthm,mathtools}
\usepackage{bm}
\usepackage{booktabs}
\usepackage{graphicx}
\usepackage{placeins}
\usepackage{cite}
\usepackage[hidelinks]{hyperref}
\title{EvoNash-MARL: A Closed-Loop Multi-Agent Reinforcement Learning Framework for Medium-Horizon Equity Allocation}
\author{
Chongliu Jia, Yi Luo$^{1}$, Sipeng Han$^{2}$, Pengwei Li$^{1}$,\\
Jie Ding$^{4}$, Youshuang Hu$^{1}$, Yimiao Qian$^{3}$, and Qiya Wang$^{1}$\\[0.4em]
\normalsize Chongliu Jia, chongliujia@gmail.com\\
\normalsize $^{1}$University of Connecticut, yi.luo@uconn.edu, parkwest.lee@gmail.com, youshuang.hu@uconn.edu, bll24001@uconn.edu\\
\normalsize $^{2}$Xi'an University of Technology, 1211913010@stu.xaut.edu.cn\\
\normalsize $^{3}$Fudan University, 24210740058@m.fudan.edu.cn\\
\normalsize $^{4}$Iowa State University, jieding@iastate.edu}
\date{}

\begin{document}
\maketitle

\begin{abstract}
Medium- to long-horizon equity allocation is challenging due to weak predictive structure, non-stationary market regimes, and the degradation of signals under realistic trading constraints. Conventional approaches often rely on single predictors or loosely coupled pipelines, which limit robustness under distributional shift.

This paper proposes EvoNash-MARL, a closed-loop framework that integrates reinforcement learning with population-based policy optimization and execution-aware selection to improve robustness in medium- to long-horizon allocation. The framework combines multi-agent policy populations, game-theoretic aggregation, and constraint-aware validation within a unified walk-forward design.

Under a 120-window walk-forward protocol, the final configuration achieves the highest robust score among internal baselines. On out-of-sample data from 2014 to 2024, it delivers a 19.6\% annualized return, compared to 11.7\% for SPY, and remains stable under extended evaluation through 2026.

While the framework demonstrates consistent performance under realistic constraints and across market settings, strong global statistical significance is not established under white's Reality Check (WRC) and SPA-lite tests. The results therefore provide evidence of improved robustness rather than definitive proof of superior market timing performance.

\end{abstract}


\section{Introduction}
\label{sec:intro_en}

Medium- and long-horizon equity allocation remains a difficult sequential decision problem.
Unlike short-horizon signal extraction, where evaluation is often concentrated on immediate directional accuracy, a deployable allocation system must remain stable across market regimes, changing factor relations, and repeated portfolio re-estimation cycles.
The difficulty is amplified by the low signal-to-noise ratio of daily equity returns: predictive structure is weak relative to realized volatility, and a signal that appears useful before trading may become unattractive once transaction costs, capacity limits, beta drift, drawdown exposure, and tail-risk constraints are considered.
These frictions are not peripheral details in medium-horizon allocation; they directly determine whether a learned policy can be used outside a backtest.
This concern is consistent with the broader financial ML literature, which emphasizes validation discipline, data snooping control, and the gap between backtest performance and deployable strategies~\cite{lopez2018advances,bailey2014probabilistic,white2000reality,hansen2005test}.

A large body of work has studied portfolio construction and financial decision-making from different angles.
Classical portfolio selection begins with mean-variance analysis~\cite{markowitz1952portfolio}, while reinforcement learning has been used to optimize trading and execution objectives directly~\cite{moody2001learning,nevmyvaka2006reinforcement}.
With deep reinforcement learning, financial applications have expanded toward automated trading, multi-asset allocation, portfolio rebalancing, market-condition-aware control, and alpha discovery~\cite{finrl2020,finrlmeta2022,wang2019alphastock,wang2021deeptrader,zhu2025alphaqcm}.
These studies show that sequential learning is a natural formulation for financial allocation.
At the same time, they also highlight persistent limitations: many systems optimize a single policy, rely on static train-test separation, or treat execution constraints and statistical validation as downstream checks rather than as part of the training and selection loop.

This mismatch is important.
In financial applications, a policy is not useful simply because it predicts returns or improves a raw Sharpe ratio in one evaluation slice.
It must also survive regime variation, avoid unstable exposure, respect feasibility constraints, and remain interpretable enough to diagnose when the learned behavior fails.
Single-agent RL is attractive because it can optimize sequential utility, but it can be brittle under noisy rewards, sparse effective feedback, and distribution shift.
A static ensemble can reduce variance, but it does not by itself provide a mechanism for strategic replacement or best-response adaptation.
Population-based and game-theoretic RL methods provide a related set of tools for maintaining strategic diversity and updating policy mixtures~\cite{lowe2017multi,lanctot2017unified,jaderberg2017population,shi2024robustmarl,qiao2024selfplay,nguyen2025ex2psro}.
Likewise, constrained and multi-objective RL provide useful foundations for reasoning about reward, risk, feasibility, and model mismatch~\cite{sun2024crmismatch,park2024maxmin}.
However, a purely game-theoretic policy pool is incomplete if the resulting mixture is selected without considering transaction costs, benchmark-relative risk, and tail constraints.

These observations motivate the central question of this paper:
can medium/long-horizon allocation be improved by treating prediction, policy aggregation, evolution, and execution-aware selection as one closed loop rather than as separate modules?
The question is non-trivial because the components can pull in different directions.
PSRO-style aggregation favors a stable mixture over policy populations; walk-forward validation favors checkpoints that generalize across time; league best-response training encourages adaptation to the current mixture; and execution-aware scaling may suppress high-conviction signals when they become too costly or risky to express.
A useful system must coordinate these forces rather than simply stacking them.

We propose EvoNash-MARL, a medium/long-horizon predictive allocation framework that couples multi-agent policy populations, PSRO-style meta-strategy updates, league best-response training, population evolution, and execution-aware checkpoint selection.
The deployable object is not a single predictor.
It is an adaptive population of policy candidates whose signals are aggregated, stress-tested, and selected under the same walk-forward protocol used for final evaluation.
The policy design separates directional information from risk control through a direction head and a risk head.
The signal layer further applies factor neutralization, nonlinear signal amplification, feature-quality weighting, and regime-aware handling.
The execution layer incorporates transaction cost, impact, capacity, beta exposure, downside risk, and tail constraints before a checkpoint is accepted.

The contribution is deliberately narrower than a new generic RL algorithm.
We do not introduce a new equilibrium solver or a new statistical test.
Instead, the paper studies how an allocation system behaves when RL-based policy search, game-theoretic aggregation, evolutionary replacement, and feasibility-aware selection are coupled inside a fixed medium-horizon walk-forward design.
This positioning matters because financial overfitting often occurs not only in the prediction model, but also in the choice of checkpoints, baselines, and evaluation windows.
Accordingly, our empirical claims are stated with care: strong return levels are not interpreted as directly deployable alpha, and statistical tests are used to separate economic evidence from global significance claims.

Empirically, the resolved v21 configuration ranks first by robust score in the 120-window protocol.
Over 2014-01-02 to 2024-01-05 OOS, its annualized return is 19.6\% versus 11.7\% for SPY, and in an extended walk-forward check through 2026-02-10 it is 20.5\% versus 13.5\%.
The same evidence also shows the limits of the current study: global significance under WRC/SPA-lite is not established, DQN baselines underperform in this medium-horizon setup, and transfer is stronger against HYG/TLT than against QQQ.
We therefore interpret the results as conditional evidence that the closed-loop training-and-selection design improves robustness under the stated protocol, not as proof of a universally dominant trading rule.

\section{Related Work}
\label{sec:related_work_en}

\noindent\textbf{Portfolio selection and financial reinforcement learning.}
Portfolio selection has long been studied through the risk-return trade-off formalized by mean-variance analysis~\cite{markowitz1952portfolio}.
Reinforcement learning later introduced a direct sequential-control view of trading, including risk-adjusted utility optimization and execution-oriented decision-making~\cite{moody2001learning,nevmyvaka2006reinforcement}.
With the development of deep RL, financial applications expanded from single-instrument trading toward automated stock trading, portfolio rebalancing, multi-asset allocation, and alpha discovery.
FinRL provides a practical deep-RL library for automated stock trading, while FinRL-Meta further emphasizes market environments, data processing, transaction costs, and benchmarking for financial RL~\cite{finrl2020,finrlmeta2022}.
AlphaStock combines an interpretable attention mechanism with a Sharpe-oriented RL objective for winner-loser portfolio selection~\cite{wang2019alphastock}.
DeepTrader incorporates market-condition embeddings to improve risk-return balancing in portfolio management~\cite{wang2021deeptrader}.
More recently, AlphaQCM frames alpha discovery as a non-stationary and reward-sparse sequential decision problem and applies distributional RL to search for synergistic formulaic alphas~\cite{zhu2025alphaqcm}.

These studies are directly relevant because they show that financial allocation is more naturally treated as a sequential control problem than as a pure supervised prediction task.
However, most existing financial RL systems emphasize one primary component at a time: a trading agent, a market environment, a reward function, or a predictive representation.
They less often study a closed loop in which policy populations, strategic aggregation, evolutionary replacement, execution constraints, and walk-forward checkpoint selection are optimized together.
Our work focuses on this integration problem rather than on replacing these prior financial RL models.

\medskip
\noindent\textbf{Multi-agent learning, game solving, and population adaptation.}
Multi-agent reinforcement learning provides a way to reduce dependence on a single learned policy by maintaining multiple interacting strategies~\cite{lowe2017multi}.
From a game-theoretic perspective, Nash equilibrium provides a classical steady-state concept~\cite{nash1950equilibrium}, while PSRO expands a policy pool through iterative best-response computation and meta-strategy updates~\cite{lanctot2017unified}.
Population Based Training shows how model parameters and hyper-parameters can be adapted jointly during training rather than fixed in advance~\cite{jaderberg2017population}.
More recent work has refined the theoretical motivation for such population-based designs.
Robust MARL studies equilibrium learning when the deployed environment may differ from the training model~\cite{shi2024robustmarl}.
Self-play theory has begun to characterize the cost of policy adaptation when updates are constrained~\cite{qiao2024selfplay}.
Recent PSRO extensions further emphasize that equilibrium selection matters, not merely equilibrium approximation~\cite{nguyen2025ex2psro}.
Survey work on self-play also highlights the practical importance of population diversity, curriculum effects, and opponent selection in RL systems~\cite{zhang2024selfplaysurvey}.

These ideas are useful for non-stationary markets because different strategies can become effective under different regimes.
A static ensemble can average model outputs, but it does not necessarily provide a principled mechanism for replacement, best-response injection, or strategic reweighting.
EvoNash-MARL therefore uses PSRO-style aggregation and league best-response training as a way to maintain a policy population whose members can be reweighted or replaced as validation evidence changes.
The goal is not to model the market as a literal two-player game, but to use game-solving machinery as a disciplined population-management tool inside a financial allocation pipeline.

\medskip
\noindent\textbf{Constraints, model mismatch, and statistical validation.}
Financial deployment differs from many simulated RL benchmarks because the training environment is only an approximation of future market conditions.
Recent constrained RL work shows that a policy satisfying constraints during training can violate them after deployment when model mismatch exists~\cite{sun2024crmismatch}.
Robust MARL makes a related point in multi-agent settings: environmental uncertainty can alter which equilibrium behavior is admissible or desirable~\cite{shi2024robustmarl}.
Multi-objective RL also provides a useful lens because trading policies must balance return, risk, stability, and feasibility rather than maximize a single scalar reward without safeguards~\cite{park2024maxmin}.

Statistical validation is equally important.
Financial ML research has repeatedly warned that raw return or Sharpe ratios can be misleading when multiple models, windows, or hyper-parameters are explored~\cite{lopez2018advances}.
We therefore use Newey-West tests~\cite{newey1987simple}, stationary bootstrap~\cite{politis1994stationary}, White Reality Check~\cite{white2000reality}, SPA~\cite{hansen2005test}, and Deflated Sharpe Ratio~\cite{bailey2014probabilistic}.
This evaluation protocol is intentionally conservative.
The reported OOS returns are treated as conditional evidence under a fixed walk-forward design, while the lack of strong WRC/SPA-lite significance is reported as a limitation rather than hidden.
Relative to prior work, the distinguishing feature of EvoNash-MARL is therefore not a single new component, but the coupling of population-based RL, game-theoretic aggregation, execution-aware utility, and statistical validation within one medium/long-horizon allocation pipeline.

\section{Method}
\label{sec:method_en}

\subsection{Problem Formulation}
\label{subsec:problem_en}
We consider daily panel data with universe size $N$ and horizon length $T$. Let
\begin{equation}
r_{i,t}=\frac{P_{i,t}}{P_{i,t-1}}-1
\end{equation}
be asset return, $b_t$ be benchmark return (default SPY), $\mathbf{x}_t\in\mathbb{R}^{d}$ be features,
$z_t\in\{\text{BULL},\text{BEAR},\text{SIDEWAYS},\text{SHOCK}\}$ be market regime, and
$s_t\in[-s_{\max},s_{\max}]$ be position signal (long-only: $[s_{\min},s_{\max}]$).

The optimization target is risk-constrained excess performance:
Let $\pi$ denote a candidate allocation policy. The optimization target is
\begin{equation}
\max_{\pi}\ 
\text{ExcessSharpe}(\pi)
-\lambda_{\text{tail}}\text{TailRisk}(\pi)
-\lambda_{\text{stab}}\text{Instability}(\pi).
\end{equation}

\subsection{Architecture Overview}
\label{subsec:overview_en}

The allocation system is organized around three interacting objects: a policy population, a meta-strategy over that population, and a validation/execution selector.
The policy population contains heterogeneous agents with different inductive biases, so the system does not rely on one model to represent all regimes.
The meta-strategy assigns mixture weights to these agents through a PSRO-style update.
The selector then evaluates the resulting signal under benchmark-relative utility, execution penalties, and validation-window stability.

This organization is motivated by the failure modes of medium-horizon trading.
A single policy can become fragile when the market regime changes; a static ensemble can average signals but cannot replace weak members; and an unconstrained trading objective can prefer checkpoints that look profitable before costs but become unattractive after beta, drawdown, or capacity constraints are applied.
EvoNash-MARL addresses these issues by keeping policy aggregation, league replacement, evolution, and execution-aware selection inside the same training loop.
Recent self-play, PSRO, robust MARL, and multi-objective RL results support the use of population-level adaptation and explicit objective trade-offs, although our implementation is an applied allocation system rather than a direct instantiation of those theoretical models~\cite{nguyen2025ex2psro,zhang2024selfplaysurvey,shi2024robustmarl,park2024maxmin}.

\begin{figure}[t]
\centering
\includegraphics[width=0.985\linewidth]{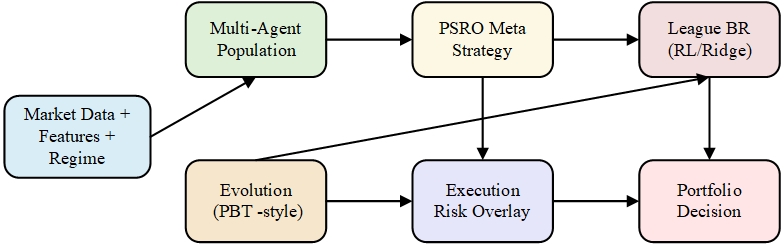}
\caption{EvoNash-MARL allocation loop. Policy populations are aggregated through a PSRO-style meta-strategy, updated through league/evolution steps, and selected through execution-aware validation.}
\label{fig:framework_en}
\end{figure}

\subsection{Feature and Regime Modeling}
\label{subsec:feature_regime_en}
Features include market momentum/volatility, cross-sectional breadth/dispersion,
tail and higher-moment descriptors, benchmark-relative features, and LS factor streams.
Regimes are identified by
\begin{equation}
\text{BULL}: \bar r_t\ge \theta_{\text{bull}},\quad
\text{BEAR}: \bar r_t\le \theta_{\text{bear}},
\end{equation}
\begin{equation}
\text{SHOCK}: \sigma_t>\kappa_{\text{shock}}\bar \sigma_t,\quad
\text{SIDEWAYS}: \text{otherwise}.
\end{equation}

\subsection{Policy Parameterization and Risk Head}
\label{subsec:policy_en}
For agent $k$, base score is
\begin{equation}
a_{k,t}=\mathbf{w}_k^\top\mathbf{x}_t+b_k+c_{k,z_t},
\end{equation}
with regime bias $c_{k,z_t}$. Base signal:
\begin{equation}
\tilde s_{k,t}=\tanh\!\left(\frac{a_{k,t}}{\max(0.15,\tau_k)}\right).
\end{equation}

If risk-head is enabled:
\begin{equation}
u_{k,t}=\sigma\!\left((\mathbf{w}_k^{(r)})^\top\mathbf{x}_t+b_k^{(r)}+c_{k,z_t}^{(r)}\right),
\end{equation}
\begin{equation}
\ell_{k,t}=\ell_k^{\min}+(\ell_k^{\max}-\ell_k^{\min})u_{k,t},
\end{equation}
\begin{equation}
s_{k,t}=\operatorname{sign}(\tilde s_{k,t})|\tilde s_{k,t}|\ell_{k,t}.
\end{equation}

\subsection{Trading PnL with Execution Constraints}
\label{subsec:pnl_en}
Let $p_t$ be executed position, $u_t=|p_t-p_{t-1}|$ be turnover. Daily PnL is
\begin{equation}
\begin{aligned}
\pi_t={}&p_{t-1}r_t-c_{\text{tc}}u_t-\lambda_{\text{risk}}|p_{t-1}|\sigma_t\\
&-\lambda_{\text{imp}}u_t^2-\lambda_{\text{cap}}|p_{t-1}|^2(1+\sigma_t),
\end{aligned}
\label{eq:pnl_en}
\end{equation}
where $c_{\text{tc}}=\text{bps}/10000$. We further apply rebalance scheduling, smoothing,
vol-targeting, and tail deleveraging overlays.

\subsection{Game Layer: PSRO Meta Strategy}
\label{subsec:psro_en}
For policy set $\{1,\dots,K\}$, payoff matrix:
\begin{equation}
A_{ij}=\bar \pi_i-\bar \pi_j.
\end{equation}
Meta strategy $\mathbf{m}\in\Delta^K$ is updated by multiplicative weights:
\begin{equation}
\mathbf{v}^{(\tau)}=A\mathbf{m}^{(\tau)},\quad
m_k^{(\tau+1)}\propto m_k^{(\tau)}\exp(\eta v_k^{(\tau)}).
\end{equation}
Nash gap:
\begin{equation}
\text{NashGap}=\max_i(A\mathbf{m})_i-\mathbf{m}^\top A\mathbf{m}.
\end{equation}
Ensemble signal:
\begin{equation}
s_t^{\text{ens}}=\sum_{k=1}^{K}m_ks_{k,t}.
\end{equation}

\subsection{Utility, Fitness, and Evolution}
\label{subsec:evo_en}
Strategy utility:
\begin{equation}
\begin{aligned}
U_k={}&\ \text{SR}(\pi_k)+\lambda_{\text{ex}}\text{SR}(\pi_k-b)\\
&-\lambda_{\text{down}}\text{DownDev}(\pi_k)-\lambda_{\text{dd}}|\text{MDD}(\pi_k)|\\
&-\lambda_{\text{cvar}}|\min(0,\text{CVaR}_\alpha(\pi_k-b))|\\
&-\lambda_{\text{worst}}|\min(0,\min_t(\pi_{k,t}-b_t))|\\
&-\lambda_{\text{con}}\cdot\text{Violation}_k.
\end{aligned}
\label{eq:utility_en}
\end{equation}

Constraint violation:
\begin{equation}
\begin{aligned}
\text{Violation}_k={}&\max(0,c^\star_{\text{cvar}}-\text{CVaR}_\alpha(\pi_k-b))\\
&+\max(0,c^\star_{\text{worst}}-\min_t(\pi_{k,t}-b_t)).
\end{aligned}
\end{equation}

Fitness:
\begin{equation}
\begin{aligned}
F_k={}&(A\mathbf{m})_k+U_k+\lambda_{\text{div}}D_k\\
&+\lambda_{\text{league}}L_k-\lambda_\beta|\beta_k-\beta^\star|,
\end{aligned}
\end{equation}
\begin{equation}
D_k=1-\frac{1}{K-1}\sum_{j\ne k}|\operatorname{Corr}(s_k,s_j)|.
\end{equation}
The resulting objective is intentionally multi-criteria.
Rather than maximizing return alone, the system trades off benchmark-relative reward, downside stability, constraint satisfaction, diversity, and exposure control.
This is consistent with recent multi-objective RL viewpoints in which the quality of a policy is not reducible to a single unconstrained reward channel~\cite{park2024maxmin}.
In our setting, this matters because a strategy with slightly lower raw return but materially better feasibility and inter-window stability is often preferable for medium/long-horizon deployment.

\subsection{League Self-Play and Best Response}
\label{subsec:league_en}
Opponent aggregate signal:
\begin{equation}
s_t^{\text{opp}}=\sum_{k}m_ks_{k,t}.
\end{equation}

Ridge BR uses $h$-day forward return:
\begin{equation}
R_t^{(h)}=\prod_{u=1}^{h}(1+r_{t+u})-1,
\end{equation}
\begin{equation}
y_t^{\text{BR}}=
\operatorname{clip}\!\left(\operatorname{sign}(R_t^{(h)})-s_t^{\text{opp}}\right),
\end{equation}
then ridge fitting and parameter blending.

RL-hybrid BR uses Q-learning with state $(\mathbf{x}_t,\text{pos}_t,s_t^{\text{opp}})$
(optionally with RFF expansion), and discrete position/leverage actions.
Reward without risk-head:
\begin{equation}
\begin{aligned}
r_t={}&\pi_t+\omega_h\left(p_tR_t^{(h)}-\Pi_{t:t+h}^{\text{opp}}\right)\\
&-\pi_{t+1}^{\text{opp}}-\lambda_{\text{pos}}|p_t|.
\end{aligned}
\end{equation}
Q-update:
\begin{equation}
\begin{aligned}
Q(s_t,a_t)\leftarrow{}&Q(s_t,a_t)\\
&+\eta\left[r_t+\gamma\max_{a'}Q(s_{t+1},a')-Q(s_t,a_t)\right].
\end{aligned}
\end{equation}

\subsection{Nonlinear Signal and Feature Quality Modules}
\label{subsec:nonlinear_en}
\textbf{Factor neutralization:}
\begin{equation}
\hat\beta=(F^\top F+\lambda I)^{-1}F^\top(s-\bar s),\quad
s^{\text{neu}}=s-\omega(F\hat\beta).
\end{equation}

\textbf{Signal amplification:}
\begin{equation}
m_t=(|s_t|-\tau)_+,\quad \tilde s_t=\operatorname{sign}(s_t)m_t^\gamma,\quad
s_t^{\text{amp}}=s_t+(g-1)\tilde s_t.
\end{equation}

\textbf{Signal quality gate:}
\begin{equation}
q_t=q_{\min}+(1-q_{\min})c_t^\nu,\quad s_t^{\text{gate}}=q_t s_t.
\end{equation}

\textbf{Feature-quality reweighting:}
\begin{equation}
Q_j=(1-\omega_r)\big(0.7C_j^{(1d)}+0.3C_j^{(h)}\big)+\omega_r C_j^{(\text{reg})},
\end{equation}
\begin{equation}
\begin{aligned}
w_j={}&(1-\alpha)+\alpha\cdot\operatorname{clip}\!\left(
\frac{Q_j}{\operatorname{median}(Q_{Q>0})+\epsilon},0.4,2.5\right).
\end{aligned}
\end{equation}

\subsection{Execution Scale Optimization}
\label{subsec:exec_opt_en}
We optimize scale $s\in[s_{\min},s_{\max}]$ by
\begin{equation}
\begin{aligned}
J(s)={}&\mathbb{E}[\pi_t(s)-b_t]\\
&-\lambda_{\text{cvar}}|\min(0,\text{CVaR}_\alpha(\pi(s)-b))|\\
&-\lambda_{\text{down}}\text{DownDev}(\pi(s)-b)\\
&-\lambda_{\text{to}}\text{Turnover}(s),
\end{aligned}
\end{equation}
\begin{equation}
s^\star=\arg\max_s J(s).
\end{equation}

\subsection{Training Algorithm}
\label{subsec:algo_en}
Training begins by constructing the feature and regime series from the panel data $\mathcal{D}$, then initializing the policy population $\mathcal{P}$ and meta strategy $\mathbf{m}$.
The optimization is organized into tournament rounds.
Within each round, the current population is evaluated on the training split to form the payoff matrix $A$, utility vector $U$, diversity term $D$, and league-advantage term $L$.
The PSRO update then revises the meta strategy, after which the mixed policy signal is passed through factor neutralization, nonlinear amplification, quality weighting, and execution-scale optimization.

Checkpoint selection is performed only on the validation split.
The selected checkpoint is the one that improves the constrained validation objective, not necessarily the one with the largest raw return.
After validation, the population is updated through elite retention and mutation.
When the league-injection rule is active, a best-response candidate is trained against the current policy mixture and inserted into the population; in the v21 configuration this candidate is the RL-hybrid BR with a risk head and regime-specific experts.
Generation-level and round-level patience rules stop the update process when validation improvement stalls.
The final output is the best validation-selected checkpoint, or a checkpoint ensemble when ensemble selection is enabled.

\subsection{Selection Criterion and Statistical Testing}
\label{subsec:select_stats_en}
Constrained validation selection metric:
\begin{equation}
\begin{aligned}
S={}&\text{ExcessSharpe}
-\lambda_1|\min(0,\text{ExcessCVaR})|\\
&-\lambda_2|\min(0,\text{ExcessWorstDay})|
-\lambda_3\cdot \text{Violation}.
\end{aligned}
\end{equation}

Walk-forward robust score (window-level excess Sharpe $\{s_w\}_{w=1}^W$):
\begin{equation}
\text{RobustScore}
=\bar s-\lambda_{\text{std}}\operatorname{Std}(s_w)
-\lambda_{\text{min}}|\min(0,\min_w s_w)|.
\end{equation}

Statistical evidence includes Newey-West, stationary bootstrap, White Reality Check, and SPA-lite:
\begin{equation}
\begin{aligned}
T_{\text{WRC}}&=\sqrt{T}\max_j\bar d_j,\\
T_{\text{SPA}}&=\max_j\max\left(0,\frac{\sqrt{T}\bar d_j}{\hat\sigma_j}\right).
\end{aligned}
\end{equation}

\section{Objective Alignment and Design Consistency}
\label{sec:theory_en}

This section does not claim a new theorem for financial predictability.
Its role is narrower: to show that several design choices in EvoNash-MARL are aligned with standard results from no-regret learning,
multi-objective selection, and constrained optimization.
The statements below should therefore be read as consistency checks for the training loop rather than as a standalone theoretical contribution.

\subsection{Setting}
Let $\mathcal{K}=\{1,\dots,K\}$ denote the current policy population and let
$A\in\mathbb{R}^{K\times K}$ be the zero-sum payoff matrix used by the PSRO-style meta update.
Signals and leverage outputs are clipped in implementation, so we assume
\begin{enumerate}
\item \textbf{Bounded payoffs:} $|A_{ij}|\le G,\ \forall i,j$.
\item \textbf{Bounded signals:} $|s_{k,t}|\le 1$ after clipping and leverage control.
\item \textbf{Bounded quality scaling:} $q_t\in[q_{\min},1]$ for some $0<q_{\min}\le 1$.
\end{enumerate}
These are implementation-level assumptions rather than market assumptions.

\subsection{Observation 1: Standard no-regret support for the meta-strategy}
Consider the multiplicative-weights update
\begin{equation}
m_k^{(\tau+1)}\propto m_k^{(\tau)}\exp(\eta v_k^{(\tau)}),\quad
v^{(\tau)}=A m^{(\tau)},
\end{equation}
with $|v_k^{(\tau)}|\le G$ and $\eta=\sqrt{2\log K/(TG^2)}$.
Then the textbook multiplicative-weights guarantee~\cite{freund1997decision} gives
\begin{equation}
\mathrm{Regret}_T\le G\sqrt{2T\log K},
\quad
\frac{\mathrm{Regret}_T}{T}\le G\sqrt{\frac{2\log K}{T}}.
\end{equation}
For zero-sum matrix games, the corresponding average Nash gap decreases at rate
$O\!\left(\sqrt{\log K/T}\right)$.

This observation is standard, but it matters here because our meta-strategy layer uses exactly this type of update.
The implication is not that the financial environment is solved by game theory; it is only that the policy-mixture update is consistent with a well-understood no-regret mechanism.

\subsection{Observation 2: the robust score favors temporal stability by construction}
Define the walk-forward robust score
\begin{equation}
R=\bar s-\lambda_{\mathrm{std}}\sigma_s-\lambda_{\mathrm{min}}|\min(0,s_{\min})|,
\end{equation}
where $\bar s$ is mean excess Sharpe across windows, $\sigma_s$ is the standard deviation of window-level excess Sharpe,
and $s_{\min}$ is the worst-window excess Sharpe.

If two strategies have the same $\bar s$ and satisfy
$\sigma_s^{(a)}\le \sigma_s^{(b)}$ together with
$| \min(0,s_{\min}^{(a)}) |\le | \min(0,s_{\min}^{(b)}) |$,
then immediately
\begin{equation}
R^{(a)}\ge R^{(b)}.
\end{equation}
Moreover, with
\begin{equation}
\begin{aligned}
\Delta \sigma&=\sigma_s^{(b)}-\sigma_s^{(a)}\ge 0,\\
\Delta d&=| \min(0,s_{\min}^{(b)}) |- | \min(0,s_{\min}^{(a)}) |\ge 0,
\end{aligned}
\end{equation}
we have
\begin{equation}
R^{(a)}-R^{(b)}\ge \lambda_{\mathrm{std}}\Delta \sigma+\lambda_{\mathrm{min}}\Delta d.
\end{equation}

This is a simple algebraic property rather than a deep theorem, but it clarifies what the model-selection rule actually rewards.
Modules such as the risk head, execution-scale optimization, and feature-quality weighting are useful only insofar as they improve these stability terms under the same mean-performance level.

\subsection{Observation 3: constrained checkpoint selection is feasible-first}
Validation selection uses
\begin{equation}
S=U-\lambda_{\mathrm{con}}V,\quad V\ge 0,
\end{equation}
where $U$ is unconstrained utility and $V$ is aggregate constraint violation.

For two checkpoints $a$ and $b$, if $U_a\ge U_b$ and $V_a\le V_b$, then
\begin{equation}
S_a\ge S_b \qquad \text{for any }\lambda_{\mathrm{con}}\ge 0.
\end{equation}
If $\lambda_{\mathrm{con}}\rightarrow\infty$, maximizing $S$ becomes equivalent to lexicographic selection:
first minimize $V$, then maximize $U$ within the minimum-violation set.

Again, the point is modest but operationally important.
The constrained selector is designed to reject checkpoints that look attractive on raw return but fail risk or feasibility requirements.
This is particularly relevant in finance, where an apparently profitable policy may be unusable once beta drift, downside risk, or scale-dependent frictions are accounted for.

\subsection{Implication for the present paper}
Taken together, the three observations explain the intended logic of the training loop.
The game layer uses a standard no-regret update to maintain a diversified policy mixture;
the robust-score objective makes temporal stability explicit rather than incidental;
and the constrained selector gives feasibility priority during checkpoint choice.
These arguments do not establish that excess return must exist in a given market.
They do show that, conditional on there being exploitable predictive structure, the optimization loop is internally coherent and pushes the system toward
low-regret aggregation, stable validation behavior, and deployable rather than purely backtest-optimal checkpoints.

This interpretation is also consistent with recent RL and MARL theory.
Adaptivity-constrained self-play~\cite{qiao2024selfplay}, robust equilibrium learning under uncertainty~\cite{shi2024robustmarl},
and max-min multi-objective RL~\cite{park2024maxmin} all emphasize that in non-stationary environments one should care not only about nominal reward,
but also about admissibility, robustness, and update discipline.
That is the sense in which the present framework is theoretically motivated.

\section{Experiments and Results}
\label{sec:exp_en}

\subsection{Experimental Setup}
\label{subsec:setup_en}
\textbf{Data and horizon:} we use a hybrid daily US-equity panel,
with the main evaluation window from 2010-01-04 to 2026-02-27.  
\textbf{Universe and features:} the resolved v21 run operates on a filtered 39-symbol liquid universe.
Features are trailing-only and include rolling market moments, volume-derived variables, benchmark-relative quantities, and four regime labels.  
\textbf{Protocol:} walk-forward evaluation is the primary protocol.
Core stability runs use 252 trading days for training, 21 for testing, step size 21, over 120 windows.
Inside each training window, the in-sample segment is split 80/20 into train and validation partitions, and checkpoint choice is frozen before the test slice is evaluated.  
\textbf{Methods and baselines:} EvoNash-MARL (v21) is compared with v17/v20b, DQN baselines,
panel-ridge baseline, plus cross-market and realistic-constraint stress tests.  
\textbf{Statistical evidence:} Newey-West, stationary bootstrap, White Reality Check (WRC), and SPA-lite.
The intended claim is economic and engineering robustness under this protocol, not strong proof of global statistical dominance.

\medskip
\noindent\textbf{Main stability, extended OOS, and statistical evidence.}
\begin{table}[!t]
\centering
\caption{Main stability results over 120 walk-forward windows (higher is better)}
\label{tab:main_120_en}
\scriptsize
\resizebox{\columnwidth}{!}{
\begin{tabular}{lrrrr}
\toprule
Model & MeanExSharpe & StdExSharpe & RobustScore & MeanBeta \\
\midrule
v21\_120w  & 0.7600 & 3.4254 & -0.0203 & 0.6731 \\
v17\_120w  & 0.7352 & 3.4361 & -0.0457 & 0.6792 \\
v20b\_120w & 0.7212 & 3.4691 & -0.0613 & 0.6651 \\
\bottomrule
\end{tabular}
}
\end{table}

Under the robust-score criterion, v21 ranks first in the 120-window evaluation.
On aligned daily OOS returns from 2014-01-02 to 2024-01-05, v21 also exceeds SPY
in absolute performance: cumulative return is 499.4\% versus 203.6\%, and
annualized return is 19.6\% versus 11.7\%.
We interpret this as economically meaningful but not decisive on its own, because the later multiple-testing results remain weak.
In particular, we do not interpret the reported return level as directly deployable alpha; under a small filtered universe, fixed walk-forward design, and model-selection loop, it should be read as conditional evidence under this protocol rather than as a live-trading claim.

\begin{figure}[!t]
\centering
\includegraphics[width=0.92\linewidth]{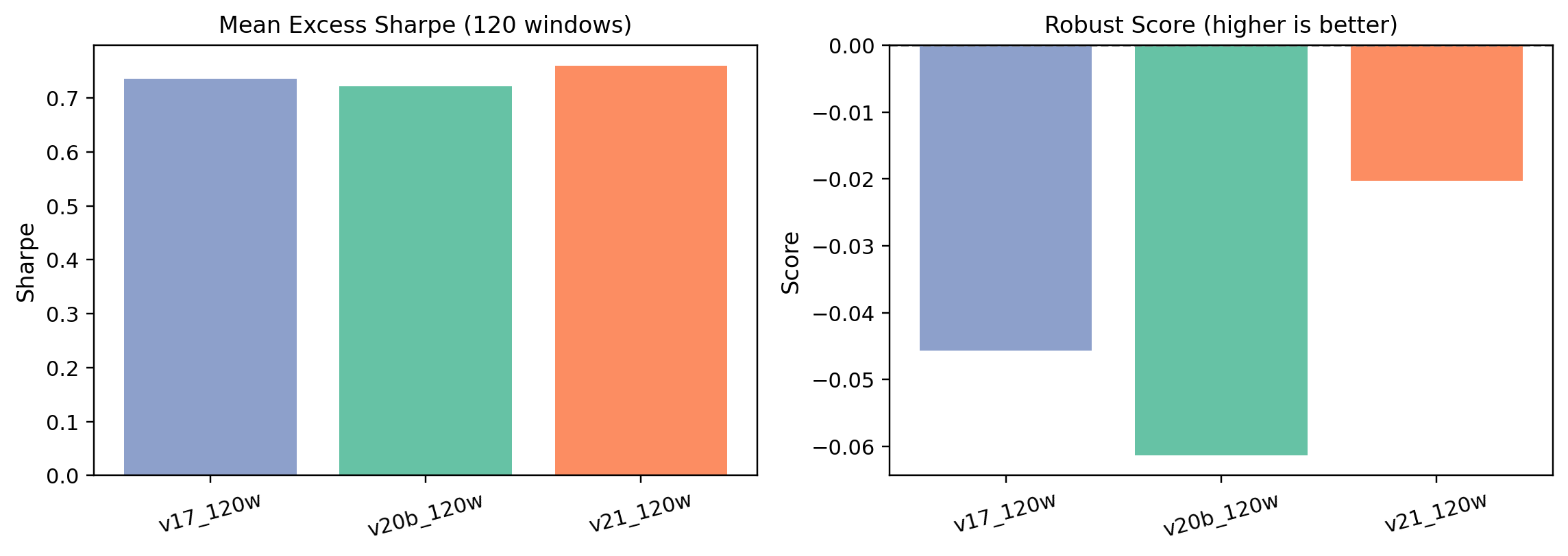}
\caption{120-window main metrics comparison (v17/v20b/v21).}
\label{fig:main120_metrics_en}
\end{figure}

\begin{figure}[!t]
\centering
\includegraphics[width=0.82\linewidth]{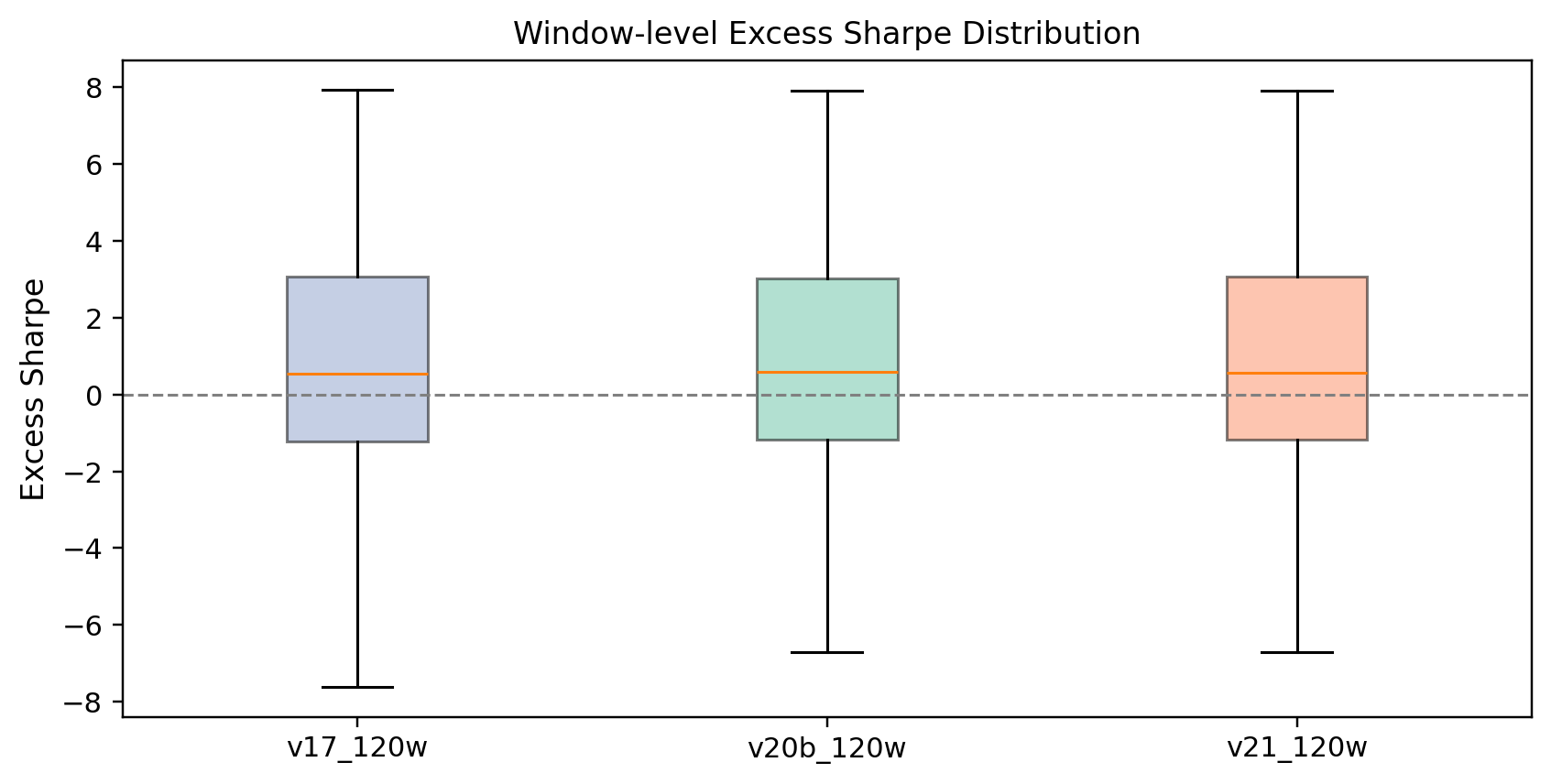}
\caption{Window-level excess Sharpe distribution.}
\label{fig:window_box_en}
\end{figure}

\begin{figure}[!t]
\centering
\includegraphics[width=0.92\linewidth]{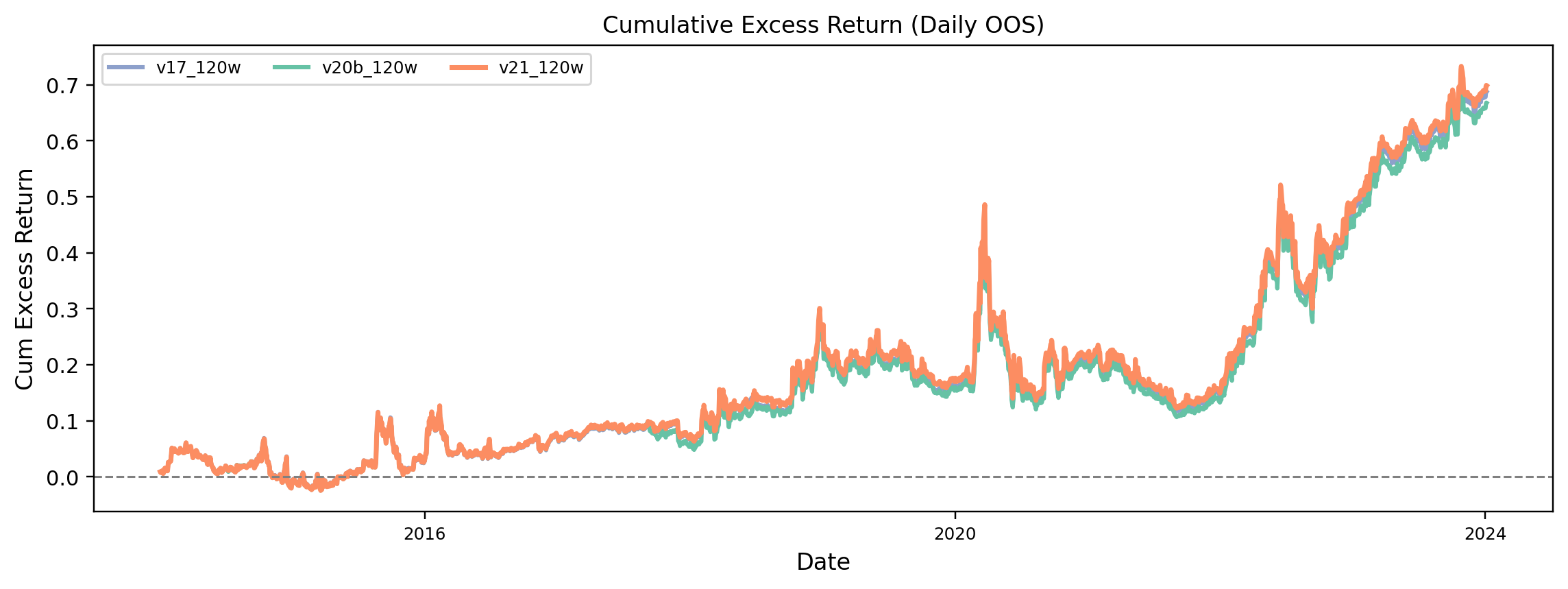}
\caption{Daily out-of-sample cumulative excess return curves.}
\label{fig:cum_excess_en}
\end{figure}

\begin{figure}[!t]
\centering
\includegraphics[width=0.92\linewidth]{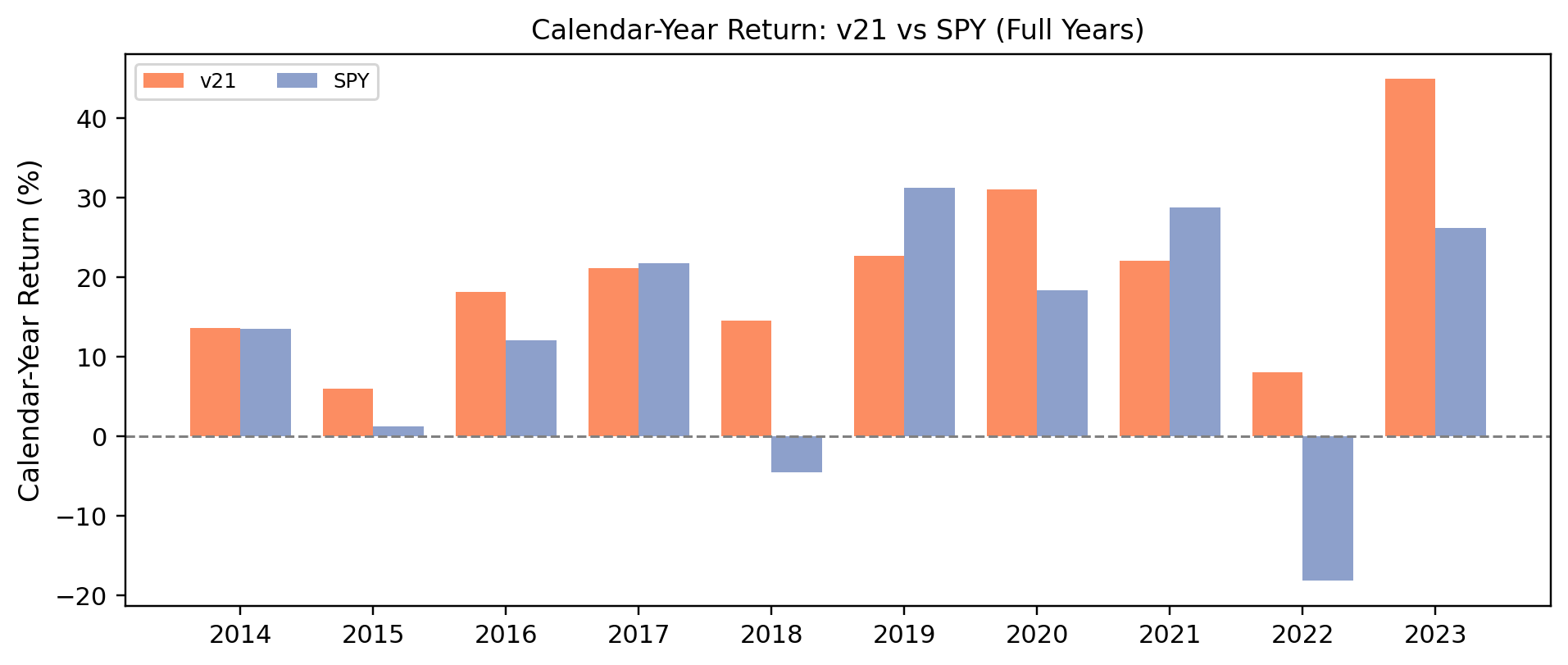}
\caption{Calendar-year return comparison between v21 and SPY (full years only).}
\label{fig:annual_vs_spy_en}
\end{figure}
Using the same walk-forward protocol (252/21/21, non-overlap) with the maximum
feasible number of windows on available data, we obtain 145 windows and daily OOS coverage
from 2014-01-02 to 2026-02-10. In this extended evaluation, v21 remains ahead
of SPY: cumulative return is 848.4\% versus 360.1\%, and annualized return is
20.5\% versus 13.5\%.

\begin{figure}[!t]
\centering
\includegraphics[width=0.92\linewidth]{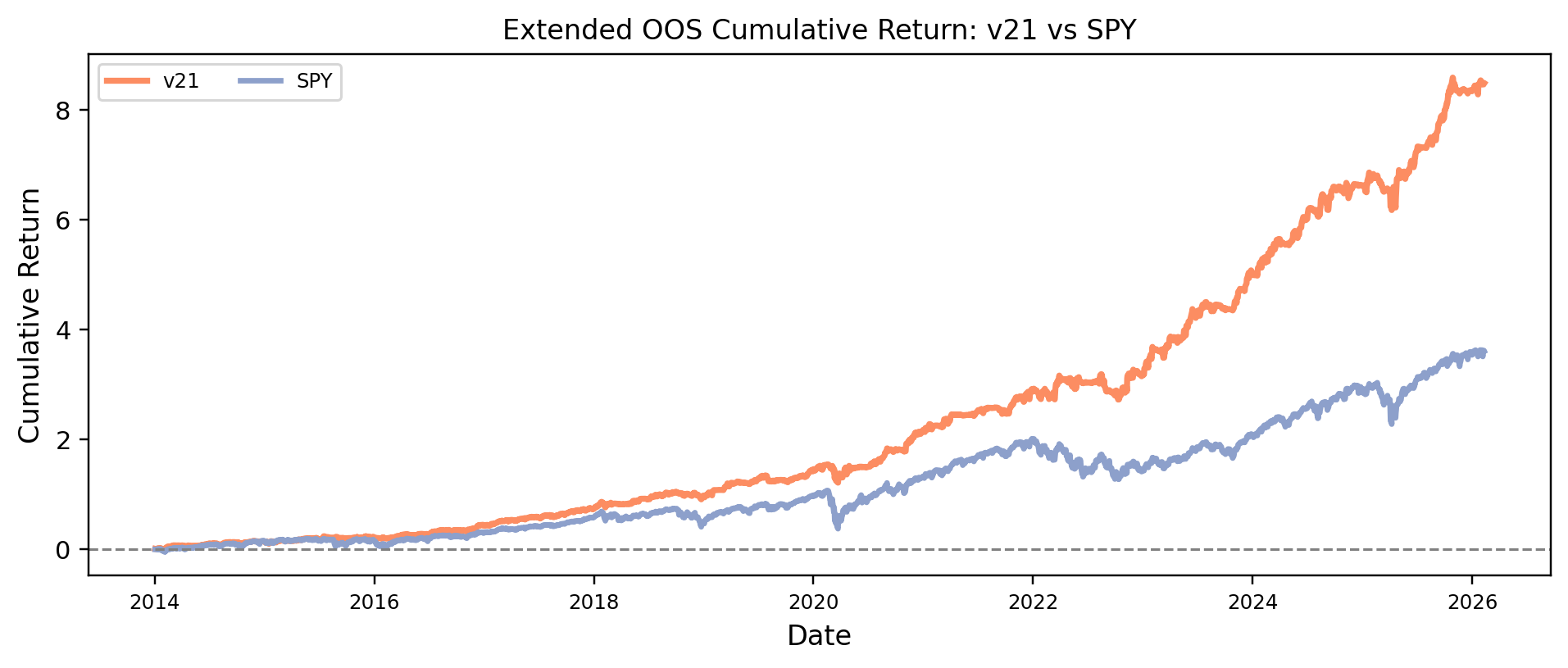}
\caption{Extended OOS cumulative return comparison (v21 vs SPY, through 2026-02-10).}
\label{fig:extended_cum_v21_spy_en}
\end{figure}
\begin{table}[!t]
\centering
\caption{Pairwise tests versus v17\_120w}
\label{tab:stats_en}
\scriptsize
\resizebox{\columnwidth}{!}{
\begin{tabular}{lrrrr}
\toprule
Candidate & MeanDiff(1d) & NW p(1-side) & SB p(1-side) & FDR-q \\
\midrule
v21\_120w  & 2.54e-06  & 0.3691 & 0.3170 & 0.6785 \\
v18\_120w  & -3.45e-06 & 0.5572 & 0.5655 & 0.6785 \\
v20b\_120w & -4.73e-06 & 0.6739 & 0.6785 & 0.6785 \\
\bottomrule
\end{tabular}
}
\end{table}

\begin{table}[!t]
\centering
\caption{Global multi-model significance tests}
\label{tab:global_stats_en}
\scriptsize
\resizebox{\columnwidth}{!}{
\begin{tabular}{lrr}
\toprule
Test & Statistic & p-value (1-side) \\
\midrule
White Reality Check & 1.2755e-04 & 0.6765 \\
SPA-lite            & 0.3475      & 0.5040 \\
\bottomrule
\end{tabular}
}
\end{table}

These tests indicate directional improvement, but they do not establish strong global statistical significance.
Accordingly, the paper does not argue that the reported return spread is sufficient by itself to prove a persistent universal alpha source.
The narrower empirical conclusion is that, under a fixed medium/long-horizon walk-forward design, the v21 training-and-selection loop produces more stable deployable checkpoints than the internal controls considered here.

\begin{figure}[!t]
\centering
\includegraphics[width=0.84\linewidth]{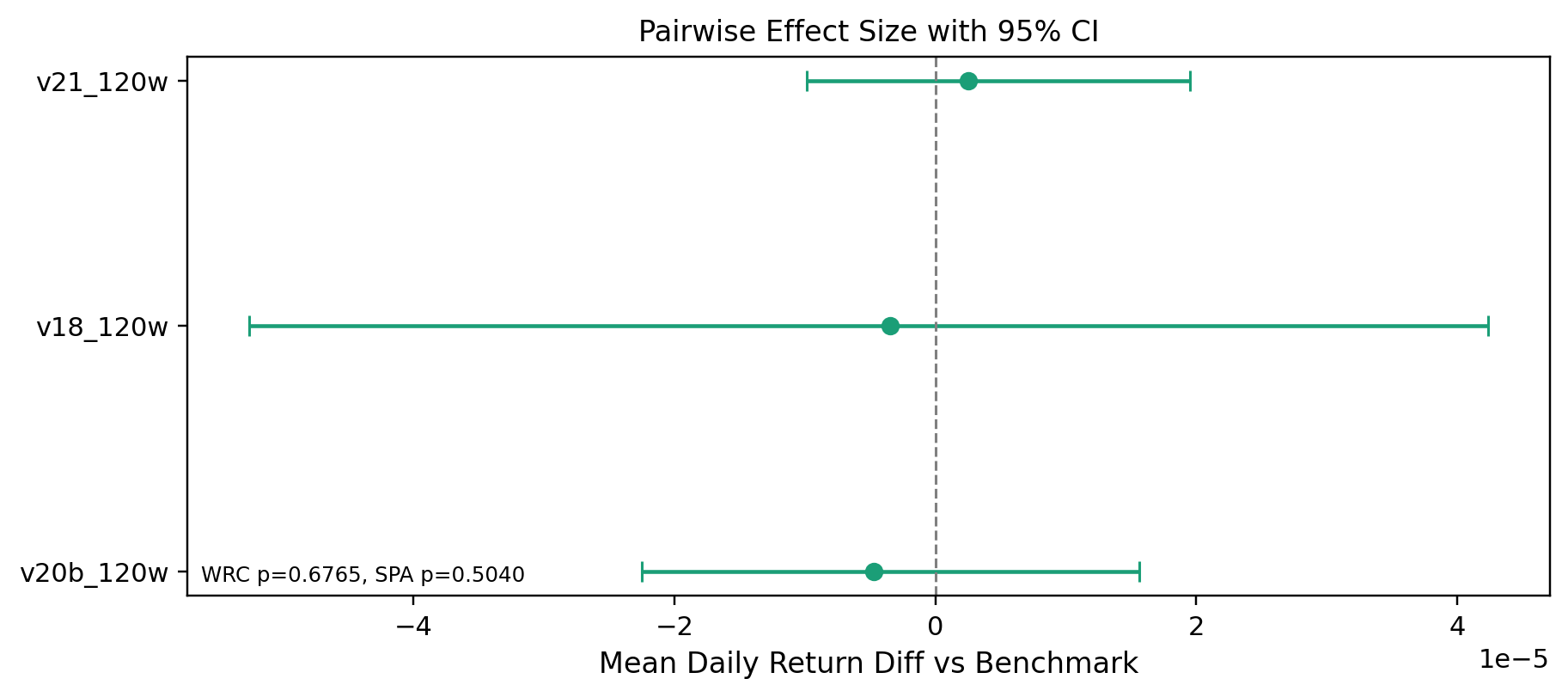}
\caption{Pairwise effect-size confidence intervals with global WRC/SPA context.}
\label{fig:sig_forest_en}
\end{figure}
\FloatBarrier

\medskip
\noindent\textbf{Baselines, transfer, and stress tests.}
\begin{table}[!t]
\centering
\caption{24-window baseline comparison}
\label{tab:baseline24_en}
\scriptsize
\resizebox{\columnwidth}{!}{
\begin{tabular}{lrrrr}
\toprule
Model & MeanExSharpe & MedianExSharpe & PosRatio & MeanBeta \\
\midrule
panel\_24w               & 2.2570  & 1.5169  & 0.7917 & 1.1212 \\
v21\_first24\_from\_120w & 0.3911  & 0.4244  & 0.5833 & 0.6024 \\
dqn\_24w\_episodes40     & -0.6031 & -0.1439 & 0.5000 & 0.5527 \\
dqn\_24w\_episodes12     & -1.0444 & 0.0883  & 0.5417 & 0.3881 \\
\bottomrule
\end{tabular}
}
\end{table}

Under this setup, DQN does not outperform the proposed method; the panel baseline is stronger but with
substantially higher beta exposure.

\begin{figure}[!t]
\centering
\includegraphics[width=0.82\linewidth]{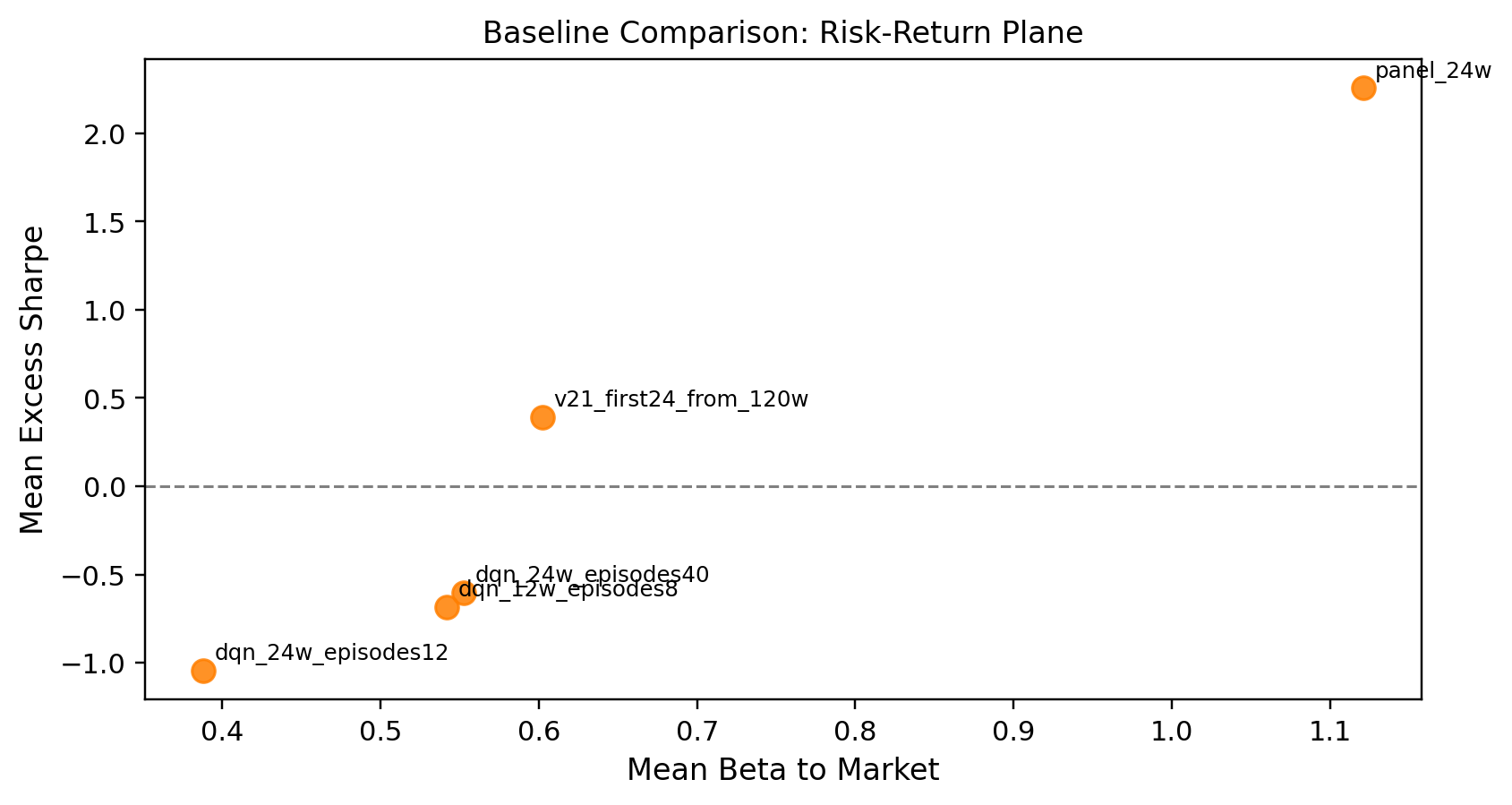}
\caption{Baseline comparison on risk-return plane (beta vs excess Sharpe).}
\label{fig:baseline_plane_en}
\end{figure}
\begin{table}[!t]
\centering
\caption{Same strategy, different benchmarks (2520 days)}
\label{tab:cross_market_en}
\scriptsize
\resizebox{\columnwidth}{!}{
\begin{tabular}{lrrr}
\toprule
Benchmark & ExcessSharpe & ExcessCumReturn & MeanExcess(1d) \\
\midrule
HYG & 1.6803 & 3.2583  & 5.9069e-04 \\
TLT & 0.7724 & 2.8636  & 6.1728e-04 \\
IWM & 0.5815 & 1.2739  & 3.7973e-04 \\
SPY & 0.5435 & 0.6987  & 2.3351e-04 \\
QQQ & 0.0195 & -0.0691 & 1.0905e-05 \\
\bottomrule
\end{tabular}
}
\end{table}

Generalization is stronger against defensive/credit-like benchmarks (HYG, TLT) and approximately neutral on QQQ.

\begin{figure}[!t]
\centering
\includegraphics[width=0.92\linewidth]{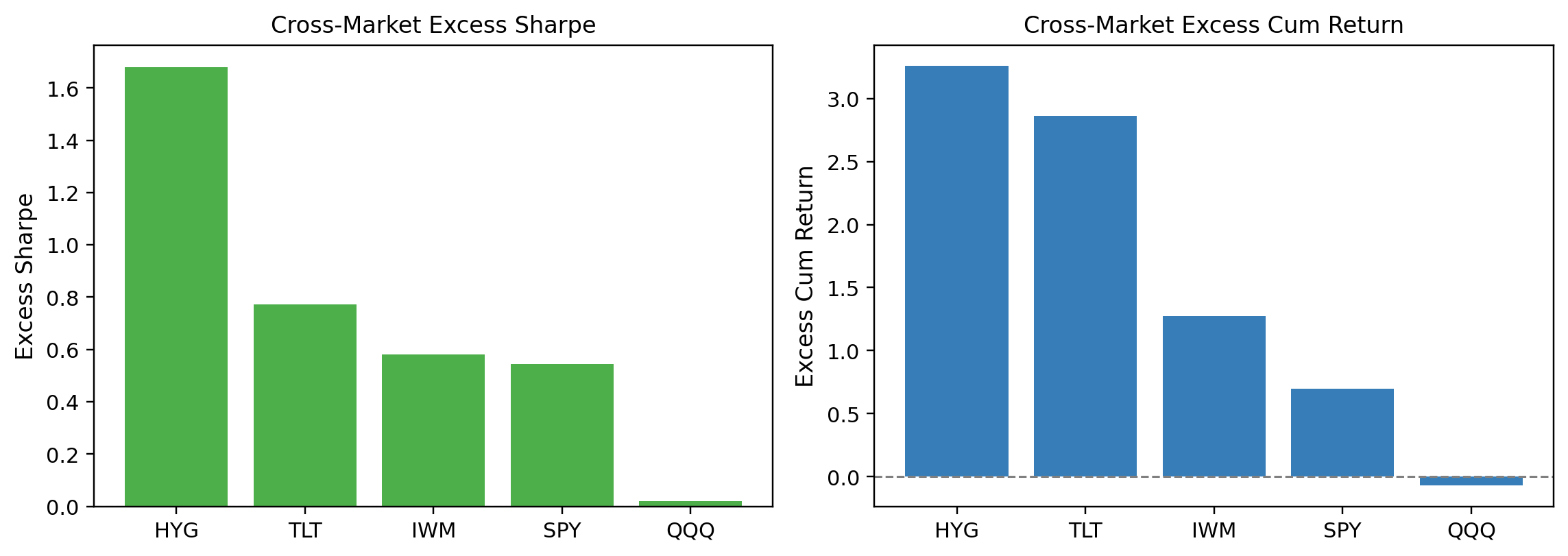}
\caption{Cross-market generalization by benchmark.}
\label{fig:cross_market_en}
\end{figure}
\begin{table}[!t]
\centering
\caption{Cost/impact/capacity stress scenarios (benchmark: SPY)}
\label{tab:stress_en}
\scriptsize
\resizebox{\columnwidth}{!}{
\begin{tabular}{lrrrr}
\toprule
Scenario & ExcessSharpe & DeltaExSharpe & ExcessCumRet & DeltaExCumRet \\
\midrule
base         & 0.5435 & 0.0000  & 0.6987 & 0.0000 \\
tc\_x3       & 0.5088 & -0.0347 & 0.6362 & -0.0625 \\
capacity\_x3 & 0.2833 & -0.2602 & 0.2818 & -0.4169 \\
all\_x2      & 0.3901 & -0.1535 & 0.4388 & -0.2599 \\
all\_x3      & 0.2366 & -0.3069 & 0.2186 & -0.4801 \\
\bottomrule
\end{tabular}
}
\end{table}

Performance degrades under stronger frictions, while excess Sharpe remains positive in all\_x3.

\begin{figure}[!t]
\centering
\includegraphics[width=0.92\linewidth]{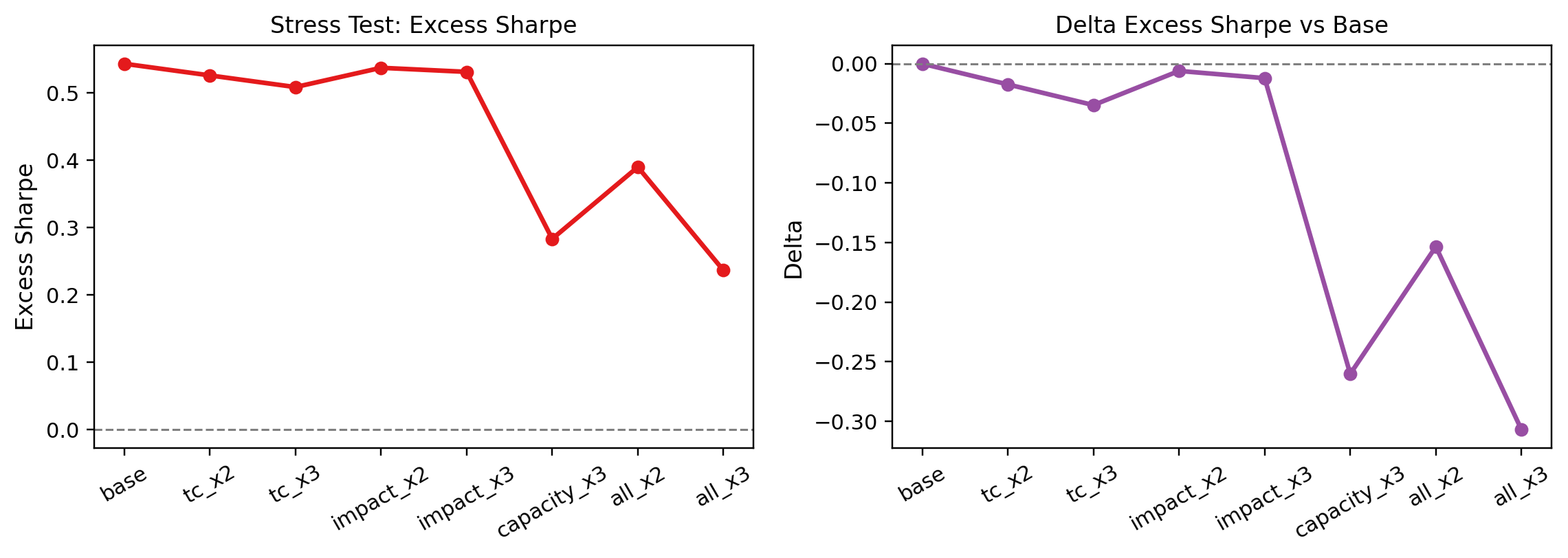}
\caption{Constraint stress sensitivity (base/tc/impact/capacity).}
\label{fig:stress_en}
\end{figure}
\FloatBarrier

\medskip
\noindent\textbf{Window-level diagnostics and ablations.}
\begin{figure*}[!t]
\centering
\begin{minipage}[t]{0.485\textwidth}
\centering
\includegraphics[width=\linewidth]{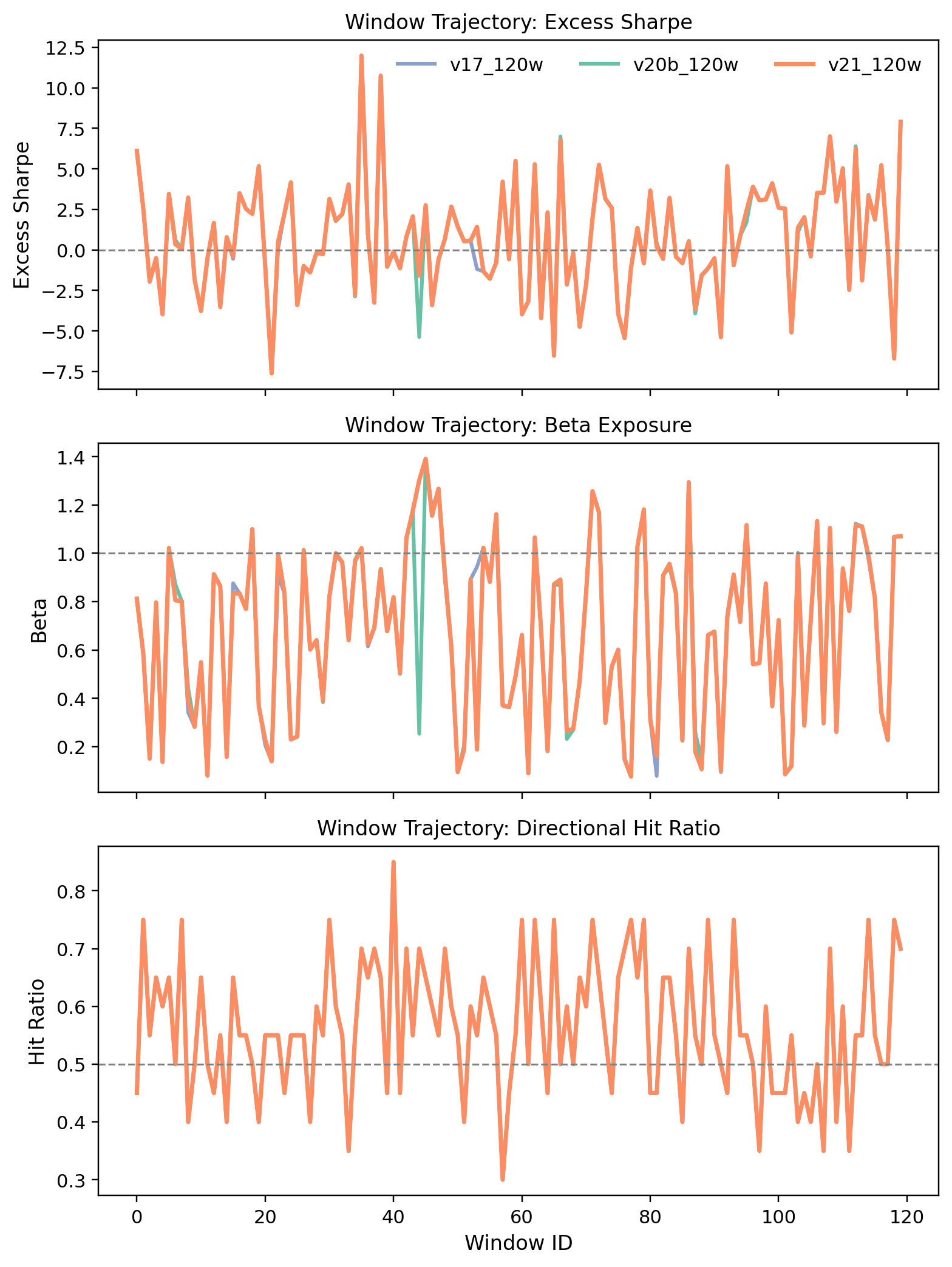}
\caption{Window-level trajectories for v17/v20b/v21 in excess Sharpe, beta exposure, and directional-hit ratio.}
\label{fig:window_diag_en}
\end{minipage}\hfill
\begin{minipage}[t]{0.485\textwidth}
\centering
\includegraphics[width=\linewidth]{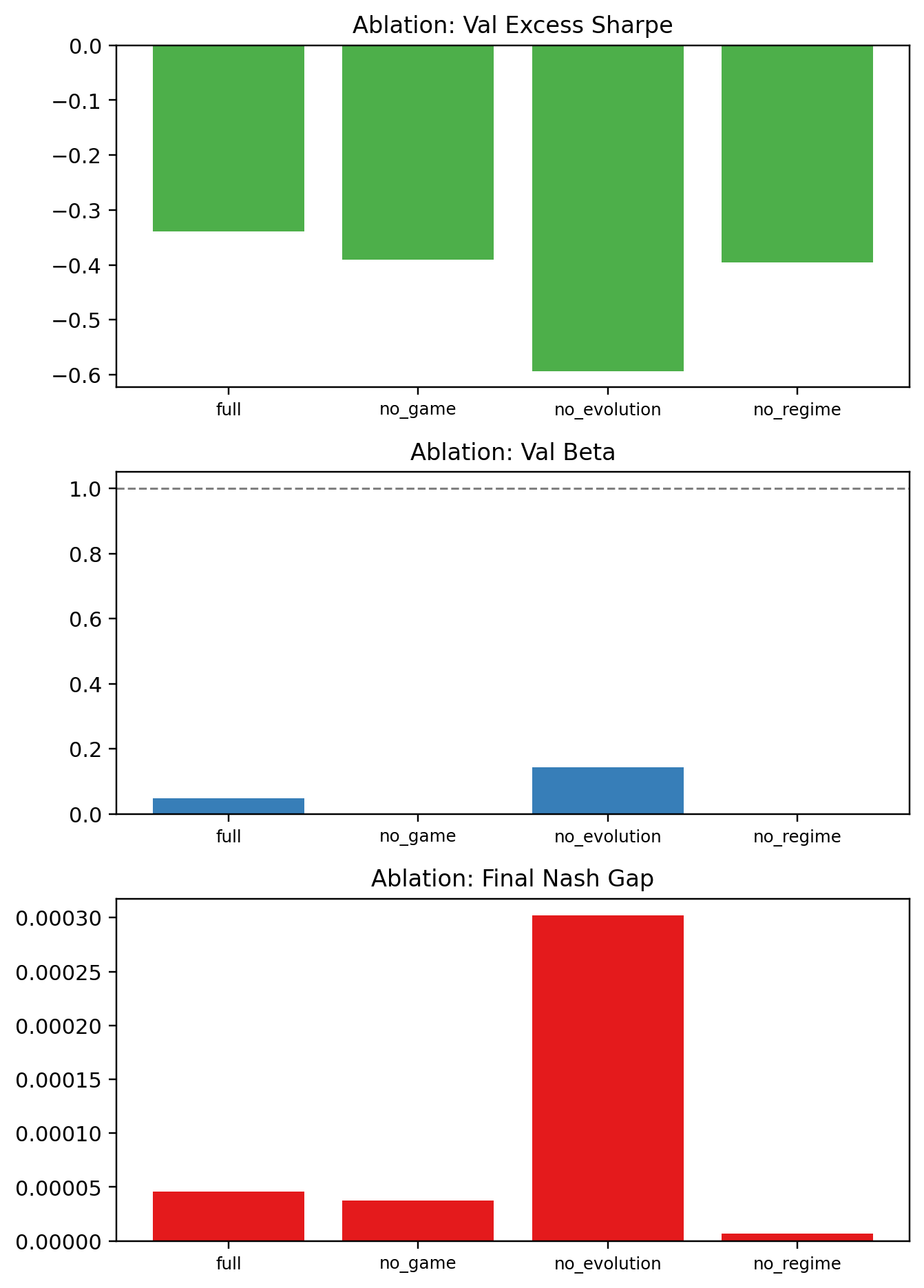}
\caption{Ablation diagnostics on validation excess Sharpe, beta, and Nash gap.}
\label{fig:ablation_en}
\end{minipage}
\end{figure*}

Figures \ref{fig:window_diag_en} and \ref{fig:ablation_en} complement the tabular results by showing
window-level stability, risk exposure variation, directional-hit behavior, and ablation diagnostics.
\FloatBarrier

\medskip
\noindent\textbf{Empirical interpretation.}
Taken together, the experiments show a consistent but bounded pattern.
The v21 configuration has the best robust score in the 120-window comparison, and the extended OOS check remains ahead of SPY through 2026-02-10.
At the same time, the global WRC/SPA-lite tests do not establish strong statistical dominance.
For this reason, the reported returns should be interpreted as conditional evidence under the fixed walk-forward protocol, not as a direct live-trading alpha estimate.

The baseline comparison clarifies where the method helps.
The DQN baselines underperform in this medium-horizon setting, which is plausible because the reward signal is noisy, delayed, and benchmark-relative.
A coarse value-based controller must learn from weak daily feedback while also aligning its actions with multi-week allocation utility.
The panel-ridge baseline is stronger in the 24-window comparison, but it also carries materially higher beta exposure.
This makes the comparison informative rather than decisive: part of the panel baseline's advantage appears to come from taking more market exposure, whereas EvoNash-MARL explicitly penalizes beta drift and selects checkpoints using constrained excess metrics.

The cross-market results suggest that the learned behavior is not a universal benchmark-independent alpha source.
Performance is stronger relative to HYG and TLT, weaker relative to QQQ, and intermediate relative to SPY and IWM.
This pattern is consistent with a risk-managed equity allocation system whose usefulness depends on the factor overlap between the strategy and the comparison benchmark.
It also reinforces the need to report transfer behavior rather than only the main SPY-relative result.

The stress tests show that capacity-related penalties are more damaging than a pure transaction-cost shock.
This is expected for medium-horizon signals: moderate linear trading costs can be absorbed when turnover is controlled, but nonlinear capacity penalties compress the strongest signals exactly when the policy wants to express them.
This observation supports including execution-scale optimization and feasibility-aware selection inside the training loop rather than applying them as final post-processing.

Overall, the results support the design under the present protocol, but stronger claims would require larger universes, stricter multiple-testing control, and additional independent OOS periods.

% Keep experiment figures/tables within this section and away from later references.
\FloatBarrier

\section{Implementation and Evaluation Protocol}
\label{sec:repro_en}

\subsection{Data and Universe}
\begin{table}[!htbp]
\centering
\caption{Raw data snapshot used in this study}
\label{tab:data_snapshot_en}
\scriptsize
\resizebox{\columnwidth}{!}{
\begin{tabular}{ll}
\toprule
Item & Value \\
\midrule
Data source & Hybrid daily US-equity panel \\
Daily observations (approx.) & 24.7 million \\
Distinct symbols (raw panel) & 6,537 \\
Raw coverage & 1962-01-02 to 2026-02-27 \\
Evaluation coverage (v21) & 2010-01-04 to 2026-02-27 \\
Working universe (v21) & 39 liquid symbols \\
\bottomrule
\end{tabular}
}
\end{table}

The starting panel is a merged daily-bar dataset assembled from long-history US-equity records and recent liquid-market coverage.
The v21 experiments do not train on the full raw universe.
Each walk-forward run applies fixed coverage and tradability filters: symbols must satisfy a minimum history threshold, pass a minimum median-price screen, and survive complete-case alignment inside the active window.
The resolved configuration caps the working universe at 39 symbols, including the main benchmark and transfer assets (SPY, QQQ, IWM, TLT, HYG, VOO, and TQQQ).
This reduced universe is intentional.
The experiment is not designed as broad cross-sectional alpha mining over thousands of names; it studies medium-horizon allocation under liquidity and execution constraints.
Daily returns are clipped at 20\% in absolute value before model training to suppress obvious bad ticks and split-adjustment artifacts.

\subsection{Walk-Forward Design}
\begin{table}[!htbp]
\centering
\caption{Primary out-of-sample protocol}
\label{tab:wf_protocol_en}
\scriptsize
\resizebox{\columnwidth}{!}{
\begin{tabular}{ll}
\toprule
Item & Value \\
\midrule
Train days per window & 252 \\
Test days per window & 21 \\
Step days & 21 \\
Maximum windows & 120 \\
Total OOS daily points & 2,520 \\
Benchmark for excess metrics & SPY \\
Model selection metric & constrained excess Sharpe \\
\bottomrule
\end{tabular}
}
\end{table}

Within each training window, features are computed only from information available up to each date.
The feature stack uses rolling market moments, volume statistics, benchmark-relative quantities, and a four-state regime label (bull, bear, sideways, shock) built from 20-day market-return and volatility summaries.
There is no future return in feature construction.
After feature construction, the in-window sample is split 80/20 into train and validation segments.
Checkpoint selection uses the constrained excess-Sharpe validation metric, and the held-out 21-day test slice is untouched until selection is complete.

\subsection{Resolved v21 Configuration}
\begin{table}[!htbp]
\centering
\caption{Main training knobs from v21 resolved configuration}
\label{tab:hp_v21_en}
\scriptsize
\resizebox{\columnwidth}{!}{
\begin{tabular}{ll}
\toprule
Parameter & Value \\
\midrule
Population size & 16 \\
Generations per round & 4 \\
Tournament rounds & 5 \\
PSRO iterations / eta & 50 / 0.25 \\
League self-play & enabled \\
BR method & RL-hybrid \\
BR RL episodes & 24 \\
Transaction cost & 3 bps \\
Impact / capacity penalty & 0.0002 / 0.0001 \\
Signal amplification & enabled ($\tau=0.1,\ \gamma=1.6$) \\
Feature-quality reweighting & enabled \\
Factor neutralization & enabled (strength 0.3) \\
Execution-scale optimization & enabled ($[0.5,1.6]$, 12 steps) \\
Beta target / penalty & 0.55 / 0.25 \\
\bottomrule
\end{tabular}
}
\end{table}

Several implementation choices are important for interpreting the results.
The system is long-only in the resolved specification, uses a 14-day rebalance interval, and optimizes against SPY as the benchmark for excess metrics.
League best-response training uses the RL-hybrid variant with 24 episodes, 21-day horizon reward shaping, regime-specific experts, and an active risk head.
Signal processing includes factor neutralization, nonlinear amplification, and feature-quality weighting.
Transaction cost, impact, capacity, beta, drawdown, and tail penalties enter either utility computation or execution-scale selection rather than being applied only after training.

\subsection{Reproduction Details}
All reported tables and figures are produced by a fixed pipeline covering statistical testing, window-stability aggregation, cross-market evaluation, stress testing, and final report assembly.
For manuscript generation, we use one frozen evidence bundle.
The manuscript numbers are therefore not hand-picked from multiple runs after the fact; once the evidence bundle is fixed, all tables and figures are generated from that bundle.
The replication package contains the statistical-test outputs, window-stability diagnostics, cross-market results, realistic-constraint stress tests, and the summary report used to build the manuscript.

\section{Conclusion, Limitations, and Future Work}
\label{sec:conclusion_en}

This paper studied EvoNash-MARL, a closed-loop framework for medium/long-horizon equity allocation.
The central design choice is to treat allocation as a population-based sequential decision problem rather than as a single-predictor forecasting task.
Within this framework, PSRO-style aggregation, league best-response training, evolutionary replacement, and execution-aware checkpoint selection are coupled under the same walk-forward protocol.

The empirical results suggest that this coupling is useful under the evaluation design considered here.
The resolved v21 configuration obtains the best robust score in the 120-window setting, remains ahead of SPY in the extended OOS check through 2026-02-10, and degrades in a plausible direction under transaction-cost, impact, and capacity stress tests.
The cross-market results are also informative: performance is stronger relative to HYG and TLT, but much weaker relative to QQQ, suggesting that the learned behavior is not a generic source of transferable alpha across all benchmarks.

The results should be interpreted with caution.
The study uses a relatively small filtered universe, and the reported return levels should not be read as directly deployable live-trading alpha.
Global strong significance is not established under the WRC/SPA-lite tests, and residual data-snooping or model-selection bias cannot be ruled out.
Execution costs, market impact, and capacity are modeled through reduced-form penalties rather than order-book simulation or broker-level execution data.
Baseline comparability is also imperfect when models carry materially different beta exposure.

Future work should therefore focus less on adding architectural components and more on stress-testing the existing loop.
Useful extensions include larger and more diverse universes, stricter multiple-testing control, causal-safe online adaptation, cross-asset evaluation, and more realistic execution modeling.
A further direction is to improve interpret-ability by decomposing strategy contributions across regimes, risk exposures, and policy subpopulations.

\FloatBarrier
\bibliographystyle{IEEEtran}
\bibliography{references_template}

\end{document}